\documentclass{article}

\usepackage{PRIMEarxiv}

\usepackage[utf8]{inputenc} 
\usepackage[T1]{fontenc}    
\usepackage{url}            
\usepackage{booktabs}       
\usepackage{amsfonts}       
\usepackage{nicefrac}       
\usepackage{microtype}      
\usepackage{lipsum}
\usepackage{fancyhdr}       
\usepackage{graphicx}       
\graphicspath{{media/}}     

\usepackage{amsmath}
\usepackage{mathtools}
\usepackage{bm}
\usepackage{subcaption}
\usepackage{booktabs}
\usepackage{multirow}
\usepackage{setspace}
\usepackage{caption}
\usepackage[flushleft]{threeparttable}
\usepackage[ruled,vlined]{algorithm2e}

\title{Dynamic Spatiotemporal Graph Convolutional Neural Networks for Traffic Data Imputation with Complex Missing Patterns
}

\author{
  Yuebing Liang \\
  Department of Urban Planning and Design \\
  The University of Hong Kong \\
  Hong Kong, China \\
  \texttt{yuebingliang@hku.hk} \\
   \And
  Zhan Zhao\thanks{Corresponding Author} \\
  Department of Urban Planning and Design \\
  The University of Hong Kong \\
  Hong Kong, China\\
  \texttt{zhanzhao@hku.hk} \\
   \And
  Lijun Sun \\
  Department of Civil Engineering \\
  McGill University \\
  Montreal, QC, Canada\\
  \texttt{lijun.sun@mcgill.ca} \\
}

\begin{document}
\maketitle

\begin{abstract}
Missing data is an inevitable and ubiquitous problem for traffic data collection in intelligent transportation systems. Despite extensive research regarding traffic data imputation, there still exist two limitations to be addressed: first, existing approaches fail to capture the complex spatiotemporal dependencies in traffic data, especially the dynamic spatial dependencies evolving with time; second, prior studies mainly focus on randomly missing patterns while other more complex missing scenarios are less discussed. To fill these research gaps, we propose a novel deep learning framework called Dynamic Spatiotemporal Graph Convolutional Neural Networks (DSTGCN) to impute missing traffic data. The model combines the recurrent architecture with graph-based convolutions to model the spatiotemporal dependencies. Moreover, we introduce a graph structure estimation technique to model the dynamic spatial dependencies from real-time traffic information and road network structure.
Extensive experiments based on two public traffic speed datasets are conducted to compare our proposed model with state-of-the-art deep learning approaches in four types of missing patterns. The results show that our proposed model outperforms existing deep learning models in all kinds of missing scenarios and the graph structure estimation technique contributes to the model performance. We further compare our proposed model with a tensor factorization model and find distinct behaviors across different model families under different training schemes and data availability.
\end{abstract}


\section{Introduction} \label{sec:intro}

Traffic data from real traffic systems plays an essential role in transportation research and applications, such as predicting traffic conditions, planning driving routes, and optimizing traffic flows. Traffic data is mainly collected from two types of sensors: stationary sensors (e.g., loop detectors) and mobile sensors (e.g., GPS probes). However, missing data is an inevitable problem for both stationary and mobile sensors. Stationary sensors can easily suffer from device problems such as detector malfunction, communication failure and power outage, whereas the data collected from mobile sensors is usually sparse with highly erratic spatial and temporal resolutions \cite{asif2013low}. 
Missing data problems seriously affect the real-time monitoring of traffic conditions and further limit other downstream applications. Therefore, how to estimate missing data, or traffic data imputation, becomes a critical issue.

Increasing attentions have been paid to the traffic data imputation problem. Early studies mainly model traffic data of each location as time series and neglect the spatial information of traffic data \cite{smith2003exploring, zhong2004estimation}. Recent research leverages local spatial information provided by neighborhood sensors/locations to improve the imputation accuracy \cite{tak2016data,lana2018imputation}. Although these studies show the effectiveness of using spatial dependence, they fail to make full use of global spatiotemporal information. Recently, matrix (or tensor) factorization methods have been introduced for traffic data imputation, and shown to be effective in retrieving correlations across different dimensions \cite{ran2016tensor,chen2019bayesian}.
However, these methods only rely on the global low-rank structure and usually do not explicitly model the underlying local consistency such as the spatial constraints of road networks and the temporal smoothness. As a result, these models may be limited in fully capturing the complexity of spatiotemporal dependencies in traffic data. With the proven success of deep learning models in a wide range of tasks, neural network-based approaches have also been adopted for the data imputation problem, including Autoencoders \cite{duan2016efficient, li2020estimation}, Recurrent Neural Networks (RNNs) \cite{berglund2015bidirectional, li2018missing} and Convolutional Neural Networks (CNNs) \cite{asadi2019convolution, benkraouda2020traffic}. Recent studies have started to employ Graph Neural Networks (GNNs) to reconstruct traffic data \cite{wu2020inductive, ye2021spatial} and demonstrate the effectiveness of GNNs in capturing spatial dependencies at the network level.
Despite extensive research to address the missing data problem, we argue that there are still two important research gaps.

First, existing approaches fail to capture the complex spatiotemporal dependencies in traffic data. Despite the promising results of GNNs, existing studies usually assume that spatial dependencies are predefined by distance and strictly unchanged across time.
However, previous studies have demonstrated that spatial dependencies are determined by not only distance but also other factors such as road grades, human mobility, etc \cite{diao2019dynamic}.  
Moreover, 
spatial dependencies on a traffic network are not constant and can change over time. 
Recently, some novel GNN-based methods have been proposed to model the true and dynamic dependencies in traffic data, but they are all developed for prediction tasks with complete data \cite{wu2019graph, zheng2020gman}. How to uncover the dynamically changing spatial and temporal patterns from incomplete and heterogeneous data to reconstruct traffic information accurately is still a challenging issue.

Second, most existing approaches are developed for random missing scenarios and might fail to provide robust results in other complex missing patterns. Random missing scenarios are based on the assumption that missing data points are completely independent of each other, but it is possible that missing values are temporally or spatially correlated. 
In such circumstances, values are missing at consecutive time intervals or spatially neighboring locations, and imputation would be more difficult as missing values have no spatially or temporally adjacent information. 

To address the issues above, we propose a novel deep learning architecture, called Dynamic Spatiotemporal Graph Convolutional Neural Networks (DSTGCN), to provide accurate and robust imputation results for different missing patterns. It consists of several spatiotemporal blocks, each comprises a bidirectional recurrent layer to capture temporal dependencies, a graph structure estimation layer to model dynamic spatial correlations and a graph convolutional layer to capture spatial dependencies. To validate the effectiveness of our model, we conduct experiments on two public datasets, one collected from a freeway network in Los Angeles, CA, and the other from an urban road network in Seattle, WA. Extensive experiments are conducted to compare our proposed model with several state-of-the-art deep learning models in four types of missing patterns with a wide range of missing ratios. The results indicate the superior performance of DSTGCN under diverse data missing scenarios. To analyze the applicability of different model families, we further compare our model with a representative tensor factorization model named BGCP \cite{chen2019bayesian} in two experiment settings, based on whether complete training data is available or not. It is found that the deep learning model shows a clear advantage when sufficient training data is available, whereas the tensor factorization method could be more suitable when training data is sparse and missing patterns are correlated.

Our main contributions are as follows: 1) we propose a novel deep learning framework to uncover the complex spatiotemporal dependencies in traffic networks and reconstruct missing traffic data accurately and robustly; 2) we design a graph structure estimation technique to estimate dynamic spatial dependencies on the network structure from real-time traffic information; 3) comprehensive experiments are conducted based on two types of real-world traffic data, and the results show that our model outperforms the state-of-the-art deep learning methods significantly under diverse missing scenarios; 4) we provide a comparative analysis of deep learning and tensor factorization models under various experiment settings and illustrate the applicability of different model families in different cases.

\section{Literature Review}

Previous studies have developed a variety of imputation methods based on different missing patterns for different types of traffic data. The performance of a method can be greatly influenced by the specific missing pattern or data type. Therefore, in this section, we first review the methods for traffic data imputation, and then summarize the different missing patterns and traffic data used in the literature. In addition, we provide a review of GCN-based models for various traffic applications to better position our paper in the field. 

\subsection{Traffic Data Imputation Methods} \label{lit:methods}
Early models for traffic data imputation mainly rely on temporal patterns and barely utilize the spatial structure of traffic data. The simplest method is Historical Average, which fills missing values based on the average values of the same time intervals in the past \cite{smith2003exploring}. \cite{ni2005markov} employed a Bayesian network to learn the probability distribution from observed data and used the best fit to impute missing data. 
\cite{qu2009ppca} introduced a technique called probabilistic principle component analysis (PPCA) which utilizes daily periodicity and interval variation of traffic data. 
Recent research incorporates spatial information into missing data reconstruction. \cite{aydilek2013hybrid} combined support vector regression (SVR) with a genetic algorithm to capture both spatial and temporal relationships in the traffic network. A modified k-nearest neighbor (KNN) method was proposed by \cite{tak2016data} which imputes missing data based on the geometry of road sections. \cite{lana2018imputation} introduced a spatial context sensing model to reconstruct traffic data using information from surrounding sensors. 
These models demonstrate that spatial information is helpful for traffic data imputation. However, they have focused on utilizing local spatial information from neighborhood locations and failed to make full use of global spatiotemporal information.

Recently, matrix (or tensor) factorization methods have been introduced for traffic data imputation, which construct traffic data as a multi-dimensional matrix and estimate a low-rank approximation of the incomplete matrix.  \cite{ran2016tensor} formulated traffic data into a 4-way tensor and employed a tensor factorization algorithm named HaLRTC to recover missing data. \cite{chen2019bayesian} extended Bayesian probabilistic matrix factorization to high-order tensors and applied it to impute incomplete traffic data. A temporal factorization framework was proposed by \cite{chen2021bayesian}, which combines low-rank matrix factorization with vector autoregressive process. 
Compared with previous models, tensor factorization is good at capturing multi-dimensional structural dependencies and thus making imputations at the network level. However, it only applies to statistical data with a low rank and needs to be learned from scratch for every new batch of incomplete data \cite{zhang2021missing}. Moreover, considering the nonlinearity and complexity of spatiotemporal dependencies in traffic data, it might be difficult for tensor factorization models to fully retrieve traffic features and provide robust imputation with diverse missing patterns and missing ratios.

With the recent advances in deep learning, a number of deep neural network models have also been developed to address the traffic data imputation problem. Compared with the tensor factorization approach, deep learning models do not make additional assumptions about data and can be pre-trained for online application when sufficient training data is provided. \cite{berglund2015bidirectional} employed bidirectional RNNs as generative models to fill in missing gaps in text data. \cite{duan2016efficient} introduced a neural network model called denoising stacked autoencoders to solve the missing data problem. Although these approaches demonstrate the effectiveness of deep learning in the field of data imputation, they barely consider spatial information. To leverage spatial correlations, \cite{li2018missing} put forward a multi-view learning method which adopts LSTM to capture temporal dependencies and SVR to capture spatial dependencies. \cite{asadi2019convolution} proposed a convolution recurrent autoencoder for missing data imputation, using multi-range CNNs to model spatial correlations. While CNNs work well for Euclidean correlations in grid-structured data (e.g., images), the non-Euclidean relationships on irregular road networks are not considered.
Recently, GNNs have shown effectiveness in embedding the graph structure of traffic systems. Based upon this, \cite{wu2020inductive} developed a Graph Convolutional Network (GCN)-based model to recover data for unobserved sensors (i.e., kriging) and represented spatial dependencies with a fixed weighted adjacency matrix predefined by distance. \cite{ye2021spatial} employed Graph Attention Networks (GATs) to adaptively learn the spatial dependencies between adjacent sensors. However, they pre-assume that spatial dependencies only exist between sensors/locations that are close in distance and fail to capture the true and dynamic relationships across the traffic network.

\subsection{Missing Patterns and Traffic Data Types in Previous Studies}
\begin{figure}[!ht]
  \centering
  \includegraphics[width=\textwidth]{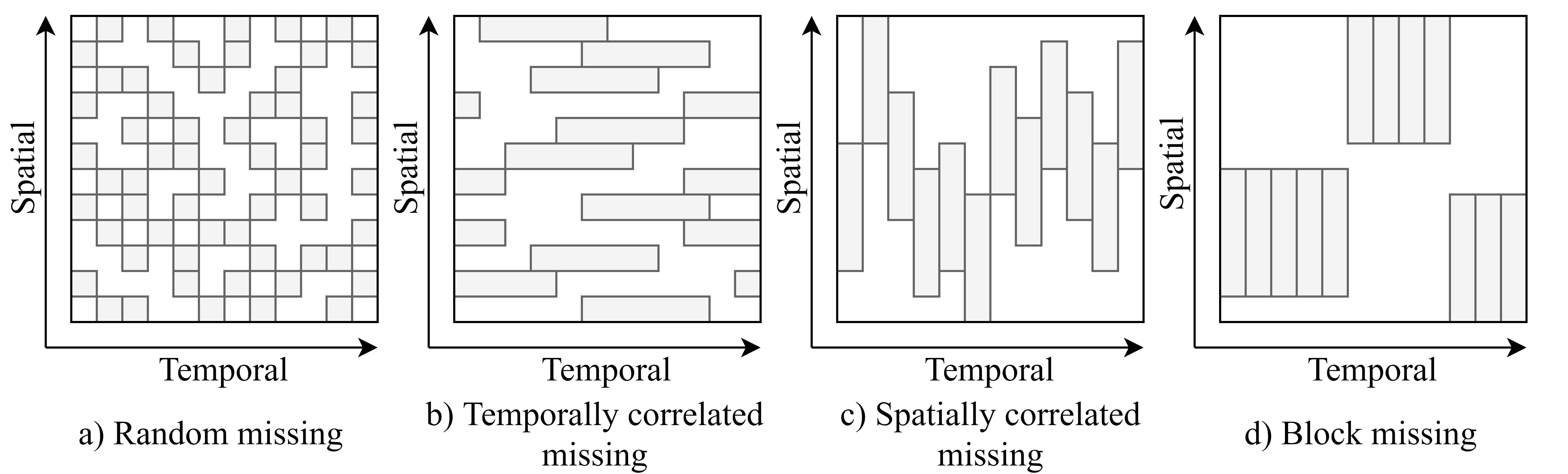}
  \caption{Patterns of missing data}\label{fig:missing_scenario}
\end{figure}

Missing patterns and data types can significantly influence the performance of a method. 
Previous studies generally classify the patterns of missing data into three classes: Missing Completely at Random, Missing at Random and Not Missing at Random \cite{little2019statistical}. Based upon this, \cite{li2018missing} classified the missing patterns in intelligent transportation systems into four classes: 1) Random Missing (RM) (Fig.~\ref{fig:missing_scenario}a), in which missing values are independent of each other; 2) Temporally correlated Missing (TCM) (Fig.~\ref{fig:missing_scenario}b), in which missing values have temporal correlations; 3) Spatially correlated Missing (SCM) (Fig.~\ref{fig:missing_scenario}c), in which missing values are related to their spatial neighboring readings; 4) Block Missing (BM) (Fig.~\ref{fig:missing_scenario}d), in which missing values are both temporally and spatially correlated and form blocks. This classification is also adopted in our research. Based upon missing ratios, missing patterns can also be classified as non-completely missing and completely missing. This study focuses on non-completely missing patterns, where at least one observed data exists in both spatial and temporal dimensions. Completely missing patterns include completely TCM \cite{wu2020inductive} and completely SCM \cite{li2020spatiotemporal}. In completely TCM, some sensors/locations are totally unobserved, whereas in completely SCM, no information is observed for some time slots. These cases are not considered in this paper.
Distinction can also be made regarding the traffic data types. Due to differences in the data collection methods and underlying road networks, traffic data can be generally classified into freeway data (FD) and urban road network data (UD). FD is usually collected with stationary sensors on freeway networks while UD is collected with mobile sensors (e.g., probe vehicles) on urban road networks. Typically, the former has higher temporal granularity, while the latter has higher spatial coverage.


Table~\ref{table:literature} summarizes the traffic data types, missing patterns and missing ratios in the literature. We can find that RM is the most commonly studied missing pattern, while the other missing patterns with temporally or spatially correlated missing values are less discussed. Compared with RM, the other patterns are more challenging due to lack of spatial or temporal adjacent information. As a result, approaches developed for random missing values might not be suitable for other missing patterns. Additionally, previous studies are usually conducted on either freeway networks or urban road networks. However, the two types of data may display different traffic characteristics due to different road design and functionality. Methods optimized for FD may not work well for UD. Missing ratios also affect the performance of models. Some previous models are developed for low missing ratios and might fail to show stable performance when the missing ratio is high. Therefore, a general approach is needed to provide accurate and robust results for different missing patterns and data types in a wide range of missing ratios. 

\begin{table}[ht!]
  \centering \footnotesize
  \caption{Methods, data and missing patterns in the literature}
  \resizebox{\linewidth}{!}{%
    \begin{tabular}{cccccccccc}
    \hline
    \multirow{2}{*}{Paper} & \multirow{2}{*}{Method} & \multicolumn{2}{c}{Data} & \multicolumn{4}{c}{Missing Patterns} &
    \multirow{2}{*}{Missing Ratio}\\
    & & FD & UD & RM & TCM & SCM & BM & & \\
    \hline
    \cite{ni2005markov} & Bayesian Networks & Y &	& Y & & & & 30\% \\
    \cite{zhang2009missing} & Least Squares SVR & & Y & Y & & & & 20-50\% \\
    \cite{qu2009ppca} & PPCA & & Y & Y & & Y & &0-50\% \\
    \cite{li2013efficient} & kernal PPCA & Y & & Y & Y& & & 5-30\% \\
    \cite{aydilek2013hybrid} &SVR+genetic algorithm& Y & & Y & & & & 1-25\% \\
    \cite{tak2016data} &modified KNN& & Y & Y & Y& Y& Y& 0.1-50\% \\
    \cite{ran2016tensor} &tensor factorization& Y & & Y & & & & 5-80\% \\
    \cite{duan2016efficient} &denoising autoencoders& Y & & Y & Y& & & 5-50\% \\
    \cite{bae2018missing} &two cokriging methods& Y & Y& Y & Y& & Y& 10-40\% \\
    \cite{lana2018imputation} &two machine learning models& Y & Y& Y & Y& & & 0-100\% \\
    \cite{li2018missing} & LSTM + SVR & Y && Y & Y& Y& Y& 5-50\% \\
    \cite{chen2019bayesian} & tensor factorization & &Y& Y & & & & 10-50\% \\
    \cite{asadi2019convolution} & recurrent convolutional autoencoders & Y& & & & & & unknown \\
    \cite{li2020estimation} & stacked autoencoders & Y& & Y& Y& & & 5-40\% \\
    \cite{li2020spatiotemporal} & prophet + Random Forest & Y& & Y& Y& & & 10-90\% \\
    \cite{wu2020inductive} & GCNs & Y& & & Y& & & 100\% \\
    \cite{chen2021bayesian} & tensor factorization & Y&Y & Y& Y& & & 30\%, 70\% \\
    \cite{zhang2021missing} &Generative Adversarial Networks& Y& & & Y& Y& Y& 10-80\% \\
    \cite{ye2021spatial} &GATs& Y& & Y& & & Y& 10-90\% \\
    \hline
    \end{tabular}}
  \label{table:literature}%
\end{table}%

\subsection{Graph Neural Networks in Traffic Research} \label{sec:GCN_review}
GNNs have been successfully applied to various prediction tasks in traffic research, including traffic speed prediction \cite{yu2017spatio}, ride-hailing demand prediction \cite{geng2019spatiotemporal} and subway passenger flow prediction \cite{ye2020multi}. To jointly extract spatiotemporal features hidden in the traffic network, researchers typically use GNNs to capture network-level spatial relations, along with RNNs or CNNs on time axis to extract temporal dependencies. By combining the recurrent architecture with diffusion graph convolution layers, \cite{li2017diffusion} introduced a deep learning framework for traffic forecasting. A purely convolutional architecture was presented by \cite{yu2017spatio} which uses graph convolutions for extracting spatial features and gated CNNs to extract temporal features. \cite{geng2019spatiotemporal} incorporated recurrent layers with a multi-graph convolution network to encode multi-level spatial correlations among regions. These approaches all extract spatial features on a fixed and pre-determined graph structure.
To uncover the true and dynamic dependencies hidden in the traffic network, \cite{wu2019graph} developed an adaptive adjacency matrix to represent hidden spatial dependencies and the matrix was learned through node embedding. \cite{diao2019dynamic} incorporated tensor decomposition into graph convolutions to estimate the change of the dependency matrix. Recently, attention mechanisms have been introduced to model spatiotemporal dependencies evolving with time. A transform attention mechanism was applied by \cite{zheng2020gman} to adaptively learn spatial and temporal dependencies from traffic features. \cite{park2020st} developed a novel spatial attention mechanism by introducing sentinel vectors to control for irrelevant features. 
However, these approaches are all developed for prediction tasks and might not be applicable to the problem of traffic data imputation. Compared with prediction tasks, the imputation problem is challenging due to the limited observed data and the diversity of missing patterns. A robust technique to model the complex spatiotemporal dependencies from incomplete and heterogeneous traffic data is still needed.

\section{Methodology} \label{sec:method}

In this section, we first introduce our problem statement and then propose a novel deep learning architecture to reconstruct missing traffic data.

\subsection{Problem Statement}
The aim of traffic data imputation is to recover complete traffic data given observed traffic data with missing values from a traffic network. A traffic network is defined as a weighted directed graph $G=(V, E, A)$, where $V$ is a set of $N = |V|$ connected nodes (i.e., sensors or road links) in $G$ and $E$ is a set of edges indicating the connectivity between nodes. $A \in \mathbb{R}^{N\times N}$ denotes a weighted adjacency matrix representing the proximity between nodes and is usually determined by a function of distance or connectivity. Each node $v$ in the traffic network records an observed value at each timestamp $t$, denoted as $x_t^v \in \mathbb{R}$. If no data is observed, $x_t^v=0$. At each time $t$, the traffic data observed on $G$ is denoted as a graph signal $X_t=\{x_t^1, x_t^2, ...x_t^N\}, X_t \in \mathbb{R}^N$. Suppose the imputation time interval is $[1, T]$, the traffic data imputation problem aims to learn a function $F(*)$  that is able to recover the complete traffic data $\overline{X}_{1:T}\in \mathbb{R}^{N\times T}$ given observed data $X_{1:T}=\{X_1, X_2, ...X_T\}, X \in \mathbb{R}^{N\times T}$ and the graph structure $G$:
\begin{equation}
    \overline{X}_{1:T}=F(X_{1:T},G).
\end{equation}

\subsection{Network Architecture}
\begin{figure}[!ht]
  \centering
  \includegraphics[width=0.5\textwidth]{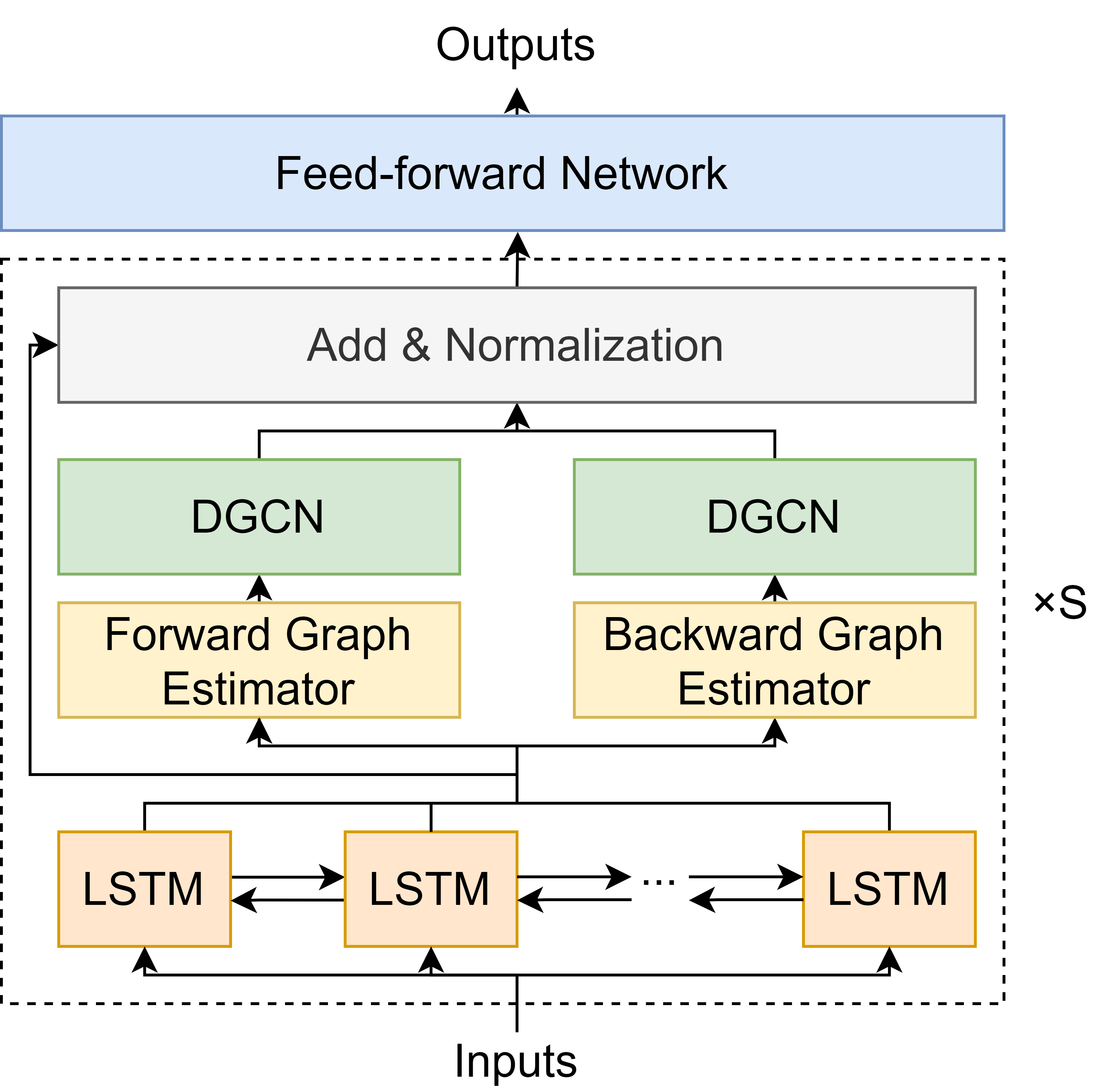}
  \caption{The architecture of DSTGCN}\label{fig:architecture}
\end{figure}

In this section, we elaborate on the architecture of our proposed model. As shown in Fig.~\ref{fig:architecture}, DSTGCN is composed of $S$ spatiotemporal blocks (ST-blocks) and an output layer. ST-blocks are used to retrieve spatiotemporal patterns from the observed traffic data. Each ST-block comprises three modules: a bidirectional recurrent layer to capture temporal features, a graph structure estimation (GSE) layer to model dynamic spatial dependencies and a graph convolutional layer to capture spatial features. The output layer is a feed-forward network which maps the output representations of the ST-blocks to the imputation results. The details of each module are described as follows.

\subsubsection{Bidirectional Recurrent Layer to Capture Temporal Features}
We leverage RNNs to capture the temporal dependencies. Specifically, we use Long Short Term Memory (LSTM) Networks , which is an efficient variant of RNNs to capture long-term and short-term dependencies in sequential data. A basic LSTM network is unidirectional, which is not suitable for the traffic data imputation problem since it can only utilize the temporal information prior to the missing time interval. To tackle this issue, we extend the unidirectional LSTM to bidirectional LSTM (BLSTM) by using two LSTM networks, one in the forward time direction and the other in the backward direction. In this way, the model is capable of exploiting information from both the past and the future \cite{berglund2015bidirectional}.

The BLSTM network is applied to each node separately and identically. For each node $v$, the input of the BLSTM layer is a sequence of feature vectors, $M^v = \{m_1^v, m_2^v,... m_T^v\}$, where $m_t^v \in \mathbb{R}^{b}$ denotes the feature vector of node $v$ at time $t$. For the BLSTM layer in the first ST-block, $m_t^v = x_t^v, m_t^v \in \mathbb{R}^{1}$. At each time step, the BLSTM layer recursively takes $m_t^v$ as the input vector and summarizes the historical and future information into two hidden state vectors $h_{f,t}, h_{b,t}$ using
\begin{gather}
\label{eq:blstm}
h_{f,t},c_{f,t} = LSTM(m_t^v, h_{f,t-1}, c_{f,t-1}), \\
h_{b,t},c_{b,t} = LSTM(m_t^v, h_{b,t+1}, c_{b,t+1}),
\end{gather}
where $h_{f,t},c_{f,t}$ are the cell memory state vector and hidden state vector in the forward direction, $h_{b,t},c_{b,t}$ are state vectors in the backward direction. To integrate the learned features from both directions, the forward and backward hidden state vectors are concatenated and mapped to the output vector ${m'}_t^v$ using a simple linear layer, given as:
\begin{equation}
\label{eq:blstm_out}
{m'}_t^v = W_l[h_{f,t}; h_{b,t}] + \beta_l ,
\end{equation}
where $W_l$ is the parameter matrix for linear transformation, $\beta_l$ is the biased term. 

\subsubsection{Graph Convolution Layer to Capture Spatial Features}
Previous research has shown that spatial dependencies of traffic data are related with directions on the traffic network \cite{liang2021vehicle}. To capture the stochastic spatial dependencies across different directions, we adopt a Difussion Graph Convolutional Network (DGCN) \cite{li2017diffusion}, which models traffic flow as a diffusion process. The working mechanism of DGCN is:
\begin{equation}
\label{eq:dgcn}
DGCN(H) = \sum_{k=1}^{K} \left(F_k(A_f) H \theta_f^k + F_k(A_b) H \theta_b^k\right),
\end{equation}
where $H$ is the input graph signals, $A_f, A_b$ are the forward and backward transition matrices respectively, $K$ is the number of diffusion steps,  $\theta_f^k, \theta_b^k$ are learning parameters of the $k$-th diffusion step that defines how the received spatial information of each node is transformed. $F_k(A)$ is a recursive function used to approximate the convolution process with $F_k(A) = 2A F_{k-1}(A) - F_{k-2}(A)$, $F_0(A) = I$, and $F_1(A) = A$.

A DGCN layer is applied to each time slot seperately and identically. For each time slot $t$, the input of the DGCN layer is a set of node features ${M'}_t = \{{m'}_t^1, {m'}_t^2,\ldots, {m'}_t^N\}$ learned from the BLSTM layer. The layer produces a new set of node features ${M''}_t = \{{m''}_t^1, {m''}_t^2,\ldots, {m''}_t^N\}$ as its output using Eq. \eqref{eq:dgcn}. Unlike the original DGCN which uses fixed transition matrices $A_f, A_b$ across different time, we generate different transition matrices $A_{f,t}, A_{b,t}$ for each time slot. Matrices $A_{f,t}, A_{b,t}$ are learned from the GSE layer, which will be introduced in the next section.

\subsubsection{GSE Layer to Model Dynamic Spatial Dependencies}
The aim of the GSE layer is to produce dynamic and directed transition matrices $A_{f,t}, A_{b,t}$ from the network typology of $G$ and real-time traffic information. Each element in the transition matrix represents the likelihood of diffusion between two nodes. Two nodes with a higher likelihood of diffusion are likely to have stronger spatial correlations. The GSE layer is composed of three steps:

\textbf{Step1:} Calculate fixed transition matrices $A_f, A_b$ from the adjacency matrix $A$ of $G$ using:
\begin{gather}
\label{eq:fix_transition}
A_f = \frac{A}{rowsum(A)},\\
A_b = \frac{A^T}{rowsum(A^T)}.
\end{gather}

\textbf{Step2:} Calculate dynamic transition matrices $\widetilde{A}_{f,t}, \widetilde{A}_{b,t}$ for each time slot using two fully-connected feed-forward networks, one for the forward direction and the other for the backward direction. The input of the feed-forward network is a set of node features $M'_t$ output from the BLSTM layer. Each feed-forward network consists of two linear transformations with a ReLU activation in between.
\begin{gather}
\label{eq:transition_dynamic}
\widetilde{A}_{f,t} = W_f^2 ReLU(W_f^1 M'_t+\beta_f^1) + \beta_f^2, \\
\widetilde{A}_{b,t} = W_b^2 ReLU(W_b^1 M'_t+\beta_b^1) + \beta_b^2, 
\end{gather}
where $W_f^1, W_b^1, W_f^2, W_b^2$ are the parameter matrices for linear transformation and $\beta_f^1, \beta_b^1, \beta_f^2, \beta_b^2$ are bias parameters.

\textbf{Step3:} There are two kinds of transition matrices for each time slot, $A_f, A_b$ from the network typology of $G$ and $\widetilde{A}_{f,t}, \widetilde{A}_{b,t}$ learned from real-time traffic information. A gated mechanism is adopted to fuse the two kinds of transition matrices. Using the forward direction as an example:
\begin{gather}
\label{eq:gate_mechanism}
g_{f,t} = sigmoid(W_g^1 A_f + W_g^2 \widetilde{A}_{f,t} + \beta_g), \\
A_{f,t} = g_{f,t} A_f + (1 - g_{f,t}) \widetilde{A}_{f,t},
\end{gather}
where $W_g^1, W_g^2$ are weight matrices for linear transformation, $\beta_g$ is the biased term.

Using $A_{f,t}, A_{b,t}$ as transition matrices, the DGCN layer updates the node features from the BLSTM layer given as:
\begin{equation}
\label{eq:dgcn_dynamic}
{M''}_t = \sum_{k=1}^{K} \left(F_k(A_{f,t}) {M'}_t \theta_f^k + F_k(A_{b,t}) {M'}_t \theta_b^k\right).
\end{equation}

\subsubsection{Residual Connection}
Training the BLSTM and DGCN layers is slow. To increase the speed of training, we employ a residual connection \cite{he2016deep} at the end of each ST-Block, followed by layer normalization to stablize the model parameters \cite{ba2016layer}:
\begin{equation}
\label{eq:st-block}
M_{out} = LayerNormalization(M_{DGCN} + M_{BLSTM}),
\end{equation}
where $M_{DGCN}, M_{BLSTM}$ are the output of the DGCN and BLSTM layer, $M_{out}$ is the output of the ST-block. 
To facilitate the residual connections, the output dimensions of the DGCN and BLSTM layer are the same. 

\subsubsection{Output Layer}
The output layer is a fully-connected feed-forward network which maps the output of the last ST-block denoted as $M_{out}^{last}$ to the imputation result $\hat{X}$. The feed-forward network consists of two linear layers and a ReLU activation in between.
\begin{equation}
\hat{X} = W_o^2 ReLU(W_o^1 M_{out}^{last}+\beta_o^1) + \beta_o^2,
\end{equation}
where $W_o^1, W_o^2$ are linear transformation matrices and $\beta_o^1, \beta_o^2$ are biased terms.

\subsection{Model Setup}
\subsubsection{Loss Function}
To make our trained model more general for all nodes in the traffic network, our loss function is defined as the reconstruction error on both observed and missing values following \cite{wu2020inductive}:
\begin{equation}
\label{eq:loss}
L(\theta) = \sum_{v=1}^{N} \sum_{t=1}^{T} (\overline{x}_t^v - \hat{x}_t^v)^2,
\end{equation}
where $\overline{x}_t^v, \hat{x}_t^v$ represent the true and predicted values of node $v$ at time $t$ respectively.

\subsubsection{Training Data Generation}
To make our model more robust to different missing ratios, we use Alg.~\ref{alg:train_sample} to generate random training samples from training data. The key idea is to randomly generate a subset of training data $X_{sample}$ and a binary mask matrix $E_{sample}$ for model training. $X_{train} = X_{sample} \odot E_{sample}$ so that missing values are masked as zeros. $\odot$ denotes the element-wise multiplication operation.
\begin{algorithm}[ht]
\SetAlgoLined
\SetKwInOut{KwIn}{Input}
\SetKwInOut{KwOut}{Output}
\KwIn{historical data $X \in \mathbb{R}^{N \times P}$ over period $[1, P]$ for training, missing type $Y$, the size of imputation window $T$, batch size $B$, training iteration $I$}
 \For{$i$ in $\{1, 2,... I\}$}{
  Randomly choose a time point $t$ within range $[1, P-T]$\;
  Obtain sampled data $X_{sample} = X_{t:t+T}$ from $X$\;
  \For{$j$ in $\{1, 2,... B\}$}{
  Randomly generate a missing ratio $r$ within range $(0, 1)$\;
  Generate a mask matrix $E_{sample}$ according to $Y$ and $r$\;
  $X_{train} = X_{sample} \odot E_{sample}$\;
  }
  Use sets $\{X_{train}^{1:B}\}$ to train DSTCGN\;
 }
 \caption{Generating training samples}\label{alg:train_sample}
\end{algorithm}

\section{Experiments} \label{sec:results}

\subsection{Data Description}
In this study, we conduct experiments on two public traffic datasets, one is collected on a freeway network and the other is collected on an urban road network. 

\textbf{METR-LA} \cite{jagadish2014big} is a traffic speed dataset \footnote{https://github.com/liyaguang/DCRNN} collected from 207 loop sensors on a freeway network of Los Angeles. The data ranges from 2012-03-01 to 2012-06-30 with a sampling rate of 5 minutes. The adjacency matrix $A$ is determined by a function of travel distance following \cite{li2017diffusion}:
\begin{equation}
A_{ij} =
    \begin{cases}
      \exp(-(\frac{dist_{ij}}{\delta})^2) & dist_{ij} \leq \kappa,\\
      0 & dist_{ij} > \kappa,
    \end{cases}   
\end{equation}
where $A_{ij}$ is the weight between sensors $i$ and $j$, $dist_{ij}$ is the distance between $i$ and $j$, $\kappa$ is the threshold and $\delta$ is a factor for normalization. In this case, $\kappa=0.1$ mile and $\delta$ is set as the standard deviation of distances.

\textbf{INRIX-SEA} \cite{cui2020graph} is a traffic speed dataset \footnote{https://github.com/zhiyongc/Graph-Markov-Network} collected from multiple data sources including GPS probes, road sensors and cell phone data on a road network in the Seattle downtown area. The data ranges from 2012-01-01 to 2012-12-31 with a sampling rate of 5 minutes. In this research, we choose a sample of road networks consisting of 223 connected road links for experiments. The adjacency matrix $A$ is provided by \cite{cui2020graph} and is a binary matrix indicating the connectivity of road links. $A_{ij} = 1$ if links $i$ and $j$ are connected and 0 otherwise.

\subsection{Missing Pattern Generation}
Following \cite{li2018missing}, we define four types of missing patterns. 
Methods of generating different missing patterns are described in \ref{missing_generation}.

\textbf{Random Missing (RM)} (Fig.~\ref{fig:missing_scenario}a): missing values are completely independent of each other and displayed as randomly scattered points for each sensor (or road). This may be due to temporary failure (e.g., power outage, communication error) for stationary sensors and the uncertainty of movement for mobile sensors. 

\textbf{Temporally Correlated Missing (TCM)} (Fig.~\ref{fig:missing_scenario}b): missing values are dependent in the time dimension and appear as a consecutive time interval for each sensor (or road). For stationary sensors, this can be caused by long-term physical damage and maintenance backlog \cite{li2018missing}. For mobile sensors this may happen when a road has no GPS probes passing by for a long time.

\textbf{Spatially Correlated Missing (SCM)} (Fig.~\ref{fig:missing_scenario}c): missing values are dependent in the spatial dimension and appear at neighboring sensors or connected road links for each time slot. For stationary sensors this may occur due to regional power outage or communication problems. For mobile sensors this may happen for urban areas with low traffic volumes.

\textbf{Block Missing (BM)} (Fig.~\ref{fig:missing_scenario}d): missing values are dependent at both spatial and temporal dimensions. In this scenario, values are missing at consecutive time intervals and spatial neighboring locations. For stationary sensors this is often caused by regional long-term malfunction. For mobile sensors this is common at mid-night when few GPS probes are working on the road network.

\subsection{Baselines}
In the numerical experiment, we compare DSTGCN with several state-of-the-art deep learning-based methods. Two groups of baselines are included. The first group contains existing imputation methods:





\textbf{Denoising Autoencoder (DAE)} \cite{duan2016efficient}: a deep learning strategy that regards the traffic states of each sensor (or road) as a vector and uses stacked DAEs to impute missing values.

\textbf{Bidirectional LSTM (BiLSTM)} \footnote{For distinction, BLSTM denotes the bidirectional recurrent layer in DSTGCN and BiLSTM denotes the bidirectional recurrent baseline model. }\cite{berglund2015bidirectional}: a recurrent architecture that predicts the missing values using LSTM model in the forward and backward time directions simultaneously.

\textbf{Convolution Bidirectional LSTM (CNN-BiLSTM)} \cite{asadi2019convolution}: a convolution recurrent autoencoder that learns the spatial information using multi-range convolutions and temporal information using bidirectional LSTM layers.


The second group contains state-of-the-art GCN-based models in traffic research. Note that these models were designed for prediction tasks and we have made every effort to adapt these baseline models for the imputation problem.

\textbf{STGCN} \cite{yu2017spatio}: a convolution framework which uses GCNs to extract spatial features and gated CNNs to extract temporal features. In this model, the adjacency matrix is considered as prior knowledge and fixed throughout training.

\textbf{GWNET} \cite{wu2019graph}: a graph neural network which captures spatial dependencies with generalized diffusion graph convolution layers and temporal dependencies with dilated convolution layers. An adaptive adjacency matrix is learned through node embedding to capture the hidden spatial dependencies in traffic data.

\textbf{GMAN} \cite{zheng2020gman}: a graph multi-attention network which uses attention mechanisms to capture spatial and temporal correlations. A spatial attention mechanism is proposed to model dynamic relevance between nodes based on real-time traffic information and graph structures.


\subsection{Experiment Settings} \label{exp: settings}
All experiments are conducted on a NVIDIA 1080 Ti GPU. We use data from the first 60\% time slots as the training set, the following 20\% as the validation set and the last 20\% as the test set. We choose $T = 72$ time steps (i.e., 5min $\times$ 72 = 6 hours) as the imputation window. 
During training, we randomly generate training samples from the training set using Alg.~\ref{alg:train_sample}; during validation and test, the imputation is conducted using a sliding-window approach on: $[t, t+T), [t+T, t+2T), [t+2T, t+3T)$, etc. The models are trained using Adam optimizer with an initial learning rate of 0.0001 and batch size of 4. Through extensive experiments, we determine the hyperparameters of our proposed model as: the number of ST-blocks $S = 2$, the diffusion step of the DGCN layer $K = 2$, the hidden state dimension of the BLSTM layer $d_h = 128$ and the output dimension of the LSTM and DGCN layer $d_o = 64$. For BiLSTM, we use the same settings as the BLSTM layer in our proposed model for fair comparison. For the other deep learning models in baselines, we use the default settings from their original proposals. 
The mean absolute error (MAE), the root-mean-squared error (RMSE) and the mean absolute percentage error (MAPE) are used to evaluate model performance:
\begin{gather}
MAE = \frac{1}{n} \sum_{i=1}^{n} |\overline{x}_i - \hat{x}_i|,\\
RMSE = \sqrt{\frac{1}{n} \sum_{i=1}^{n} (\overline{x}_i - \hat{x}_i)^2},\\
MAPE = \frac{1}{n} \sum_{i=1}^{n} |\frac{\overline{x}_i - \hat{x}_i}{\overline{x}_i}|,
\end{gather}
where $n$ is the number of missing values in the original data, $\overline{x}_i, \hat{x}_i$ are the true and predicted values for the $i$-th missing point respectively.

\subsection{Performance Analysis}

\begin{figure}[ht!]
  \centering
  \includegraphics[width=0.98\textwidth]{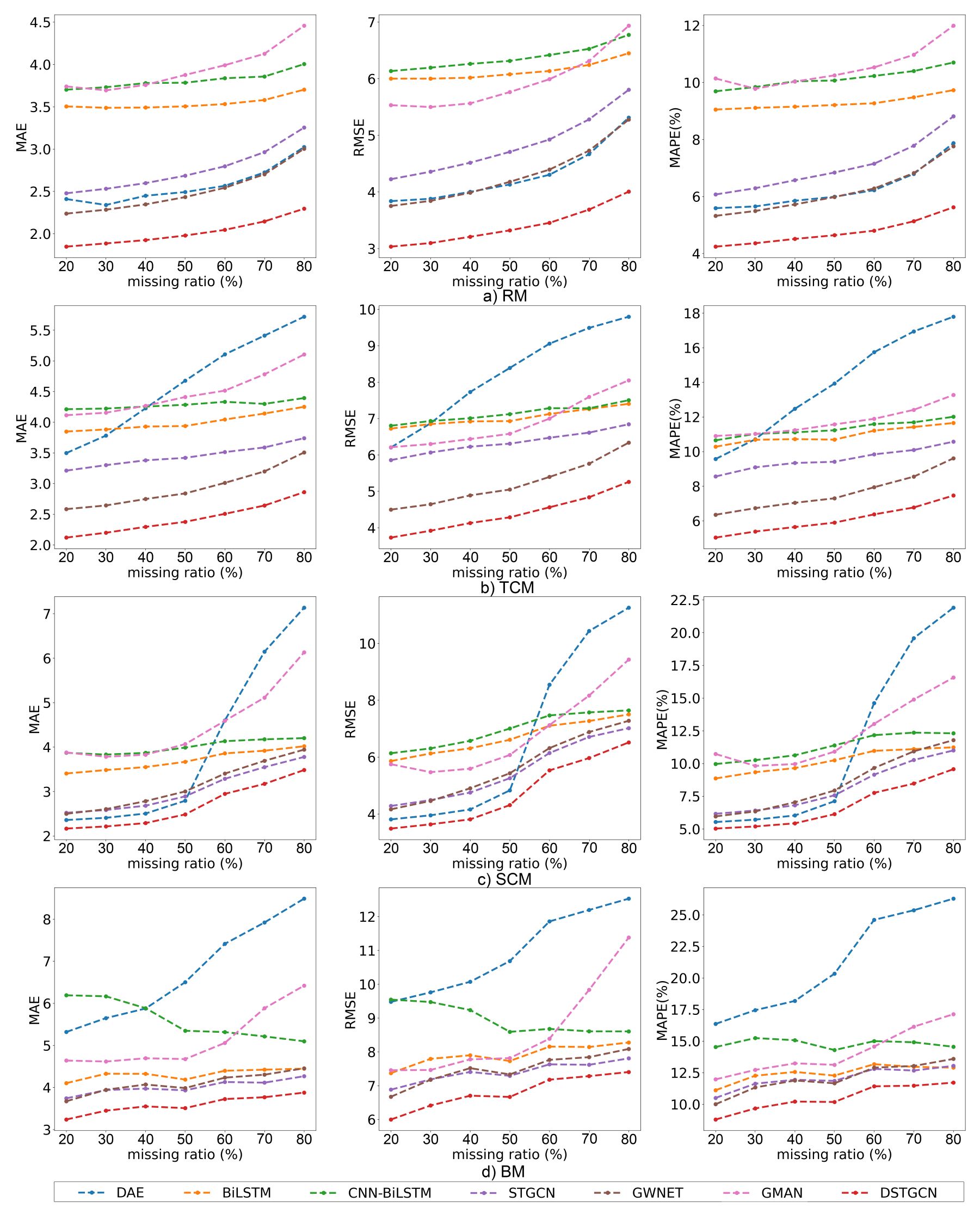}
  \caption{Performance comparison for different missing patterns on METR-LA data}\label{fig:metr_res}
\end{figure}

\begin{figure}[ht!]
  \centering
  \includegraphics[width=0.98\textwidth]{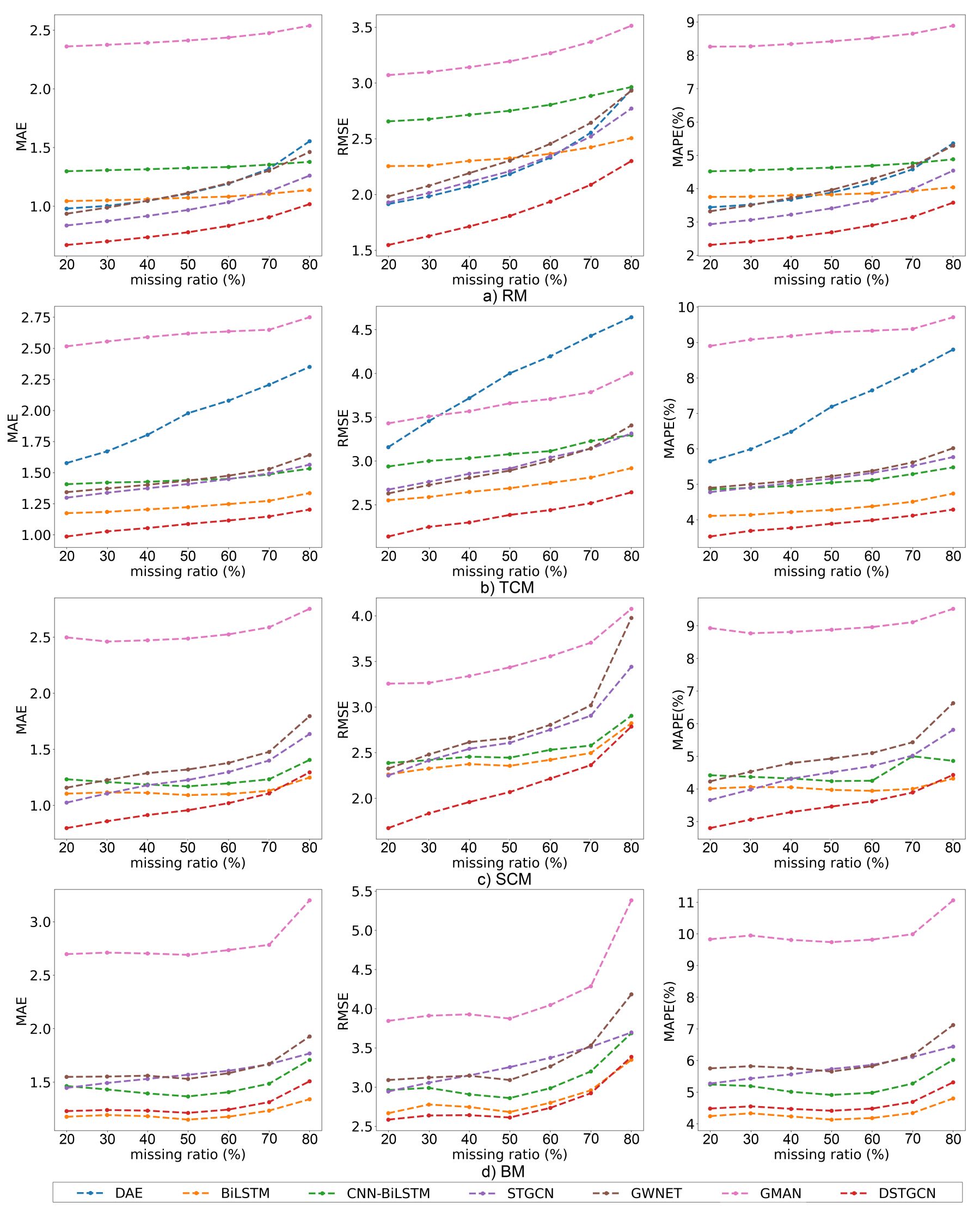}
  \caption{Performance comparison for different missing patterns on INRIX-SEA data (The results of DAE for missing patterns SCM and BM on INRIX-SEA data is presented seperately in \ref{DAE_performance})}\label{fig:inrix_res}
\end{figure}

\begin{figure}[!ht]
  \centering
  \includegraphics[width=\textwidth]{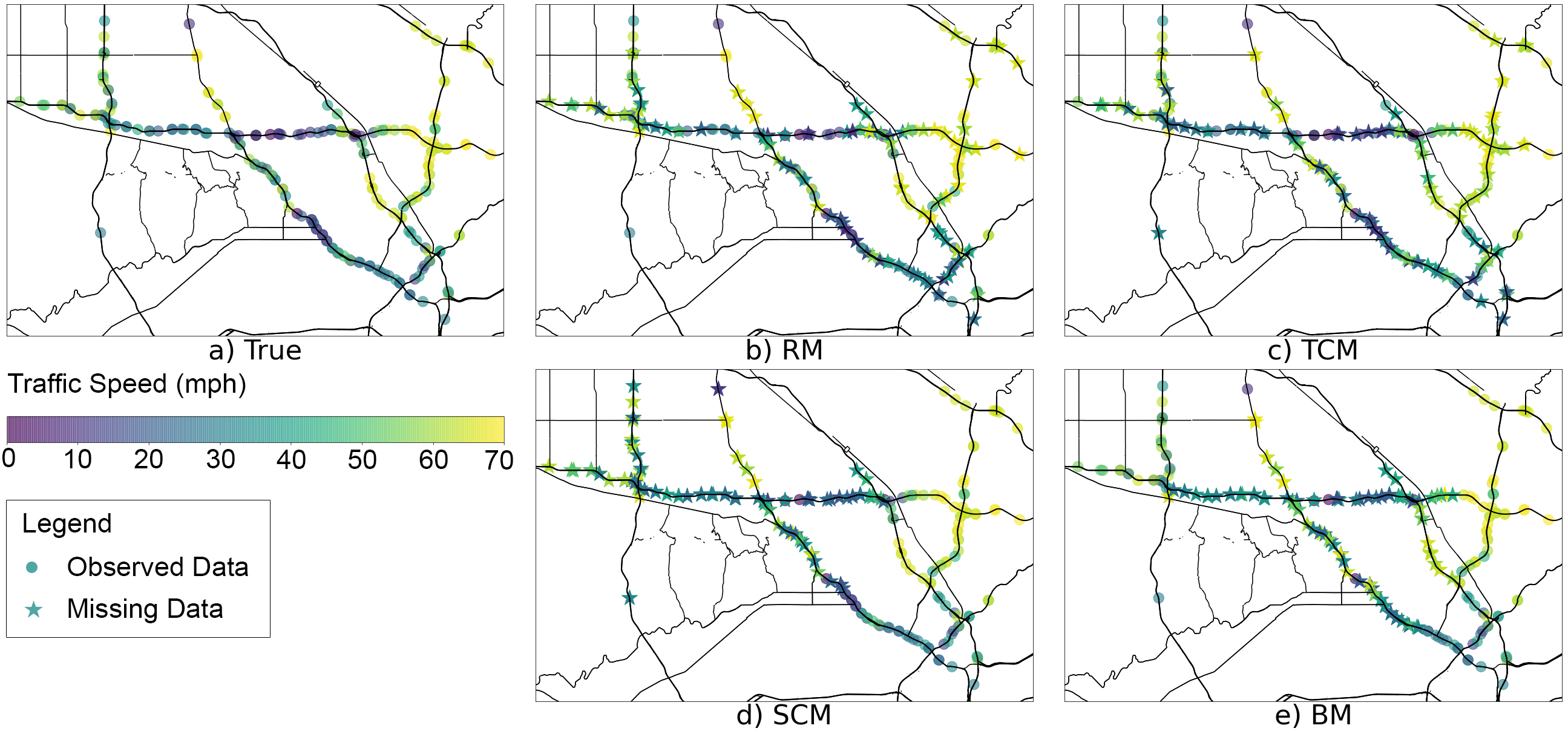}
  \caption{The imputation result of the METR-LA data at 18:00 on 2012-05-24 using DSTGCN }\label{fig:res_visual}
\end{figure}

In this section, we compare the performance of DSTGCN with the baseline models for different missing patterns and missing ratios ranging from 20\% to 80\%. Fig.~\ref{fig:metr_res} and Fig.~\ref{fig:inrix_res} display the imputation performance of different models on METR-LA data and INRIX-SEA data respectively. 
We can find that DSTGCN achieves notably superior results to the baseline models in most missing scenarios. For the METR-LA data, DSTGCN performs better than the baseline methods by a large margin for all missing patterns and missing ratios. For the INRIX-SEA dataset, DSTGCN achieves significant improvement in the imputation performance for missing patterns RM and TCM. For SCM and BM, DSTGCN outperforms existing methods with missing ratios ranging from 20\% to 70\% and shows similarly competitive performance with BiLSTM when the missing ratio reaches 80\%. This indicates that DSTGCN can provide more accurate and robust results than existing methods for diverse combinations of missing patterns and traffic data types. This is potentially because compared with the baseline models, our model can better capture the complex spatiotemporal correlations hidden in traffic data. In our model, the nonlinear temporal relationship is captured with bidirectional recurrence and the network-level spatial correlations is captured with directed graph convolutions. We also take advantage of the graph structure estimation layer which effectively models the dynamic spatial dependencies evolving over time. It should be noted that for SCM and BM on INRIX-SEA data, the advantage of DSTGCN over the baseline models (i.e., BiLSTM) gradually decreases as the missing ratio increases. This is potentially because in urban road networks, the prevalence of traffic signals tend to have a regulating effect on the spatial correlations among nearby road links. Compared with freeway networks, it is generally more difficult for GCNs to capture spatial information across urban road networks, especially when large-scale spatially adjacent information is missing.
Fig.~\ref{fig:res_visual} provides an example of the imputation result of DSTGCN in different missing scenarios with the missing ratio of 50\%. It can be seen that DSTGCN is capable of reconstructing missing values in all kinds of missing patterns. 

Apart from our proposed model, we also examine the performance of baseline models in different missing patterns, which has not been fully discussed in literature.

\textbf{DAE}: Although DAE performs well in RM, in the other missing patterns, its performance degrades tremendously with the increase of missing ratios. The poor performance of DAE in the TCM scenario is potentially because DAE regards the data of each node as a vector, which is suitable to repair isolated missing data points but fails to recover consecutive missing intervals. Additionally, DAE barely leverages spatial information, thus performs poorly in SCE and BE scenarios.

\textbf{BiLSTM}: Although BiLSTM provides competitive results on INRIX-SEA data, it performs relatively poorly on METR-LA data. This suggests that traffic conditions on urban road networks might display strong temporal dependencies, whereas traffic conditions on freeway networks have lower regularities on time axis. Compared with the other methods, BiLSTM provides the most stable performance over varying missing ratios, showing the robustness of bidirectional recurrent networks for data imputation.

\textbf{CNN-BiLSTM}: Similar to BiLSTM, CNN-BiLSTM shows stable performance for all kinds of missing patterns with a wide range of missing ratios. However, the performance of CNN-BiLSTM is consistently inferior to BiLSTM. This indicates that CNNs cannot capture the spatial dependencies on traffic networks effectively and even have a negative effect on the model performance.


\textbf{STGCN}: STGCN performs well on METR-LA data, but it fails to provide satisfying results for INRIX-SEA data. This can be explained that STGCN relies on a fixed adjacency matrix pre-determined by road connectivity to model spatial correlations and fails to capture the true spatial dependencies in urban road networks. For METR-LA data, STGCN performs relatively poorly for TCM and BM, indicating that convolution layers fail to extract temporal information effectively when missing data is temporally correlated.


\textbf{GWNET}: For METR-LA data, GWNET shows better performance than STGCN in the RM and TCM scenarios, but it performs relatively poorly for SCM and BM. This suggests that the adjacency matrix learned through node embedding fails to work well in spatially correlated scenarios. For INRIX-SEA data, its performance is inferior to STGCN in most cases, showing that the adaptive adjacency matrix is not applicable to urban road networks.

\textbf{GMAN}: Although GMAN was reported to outperform STGCN and GWNET in the task of traffic prediction, its performance is worse than STGCN and GWNET in all kinds of missing patterns for both data sets. This indicates that although the attention mechanism is capable of modeling spatiotemporal dependencies of complete traffic data, it fails to provide robust results for incomplete and heterogeneous traffic data.

In summary, none of the baseline models can provide competitive results for different missing patterns in both freeway and urban road networks. DAE fails to work well for spatially correlated missing patterns whereas STGCN cannot achieve superior results when missing values are temporally correlated. For different data types, BiLSTM and CNN-BiLSTM perform relatively poorly for freeway networks, whereas existing GCN-based models, including STGCN, GWNET and GMAN, fail to provide competitive results for urban road networks. This can be explained by the heterophily of urban road networks, where connected road segments have different traffic features due to the regulating effect of traffic lights. Research has shown that existing GNNs usually assume strong homophily and fail to generalize to heterophilous networks \cite{zhu2020beyond}.
Compared with STGCN, GWNET and GMAN fail to provide consistent improvement across different missing scenarios, suggesting that existing graph structure estimation techniques might not perform well for the imputation problem with incomplete and heterogeneous data.
Relative to the baseline, our proposed model achieves high accuracy across different missing scenarios for both freeway and urban road networks. This validates the robustness of our proposed model in capturing the complex spatiotemporal dependencies with limited observed data in both homophilous and heterophilous networks.

\subsection{Ablation Analysis}
This section aims to conduct extensive ablation studies to disentangle the contributions of different components in the proposed model. Since RM is the most commonly studied missing type in literature, we use the RM scenario for experiments. Different components of our proposed model are dropped to construct variants, which will be tested and compared against the full model. The components are listed as follows:

\textbf{BLSTM Layer}: The BLSTM layer is used to extract temporal correlations. With the BLSTM layer ablated, it is replaced with a linear layer.

\textbf{GSE Layer}: The GSE layer is used to model the dynamic spatial dependencies evolving with time. With the GSE layer ablated, we use fixed transition matrices $A_f, A_b$ to represent spatial dependencies across different time.


\textbf{DGCN Layer}: The DGCN layer is used to  extract spatial correlations. With the DGCN layer ablated, it is replaced with a linear layer.
\begin{figure}[ht!]
  \centering
  \includegraphics[width=\textwidth]{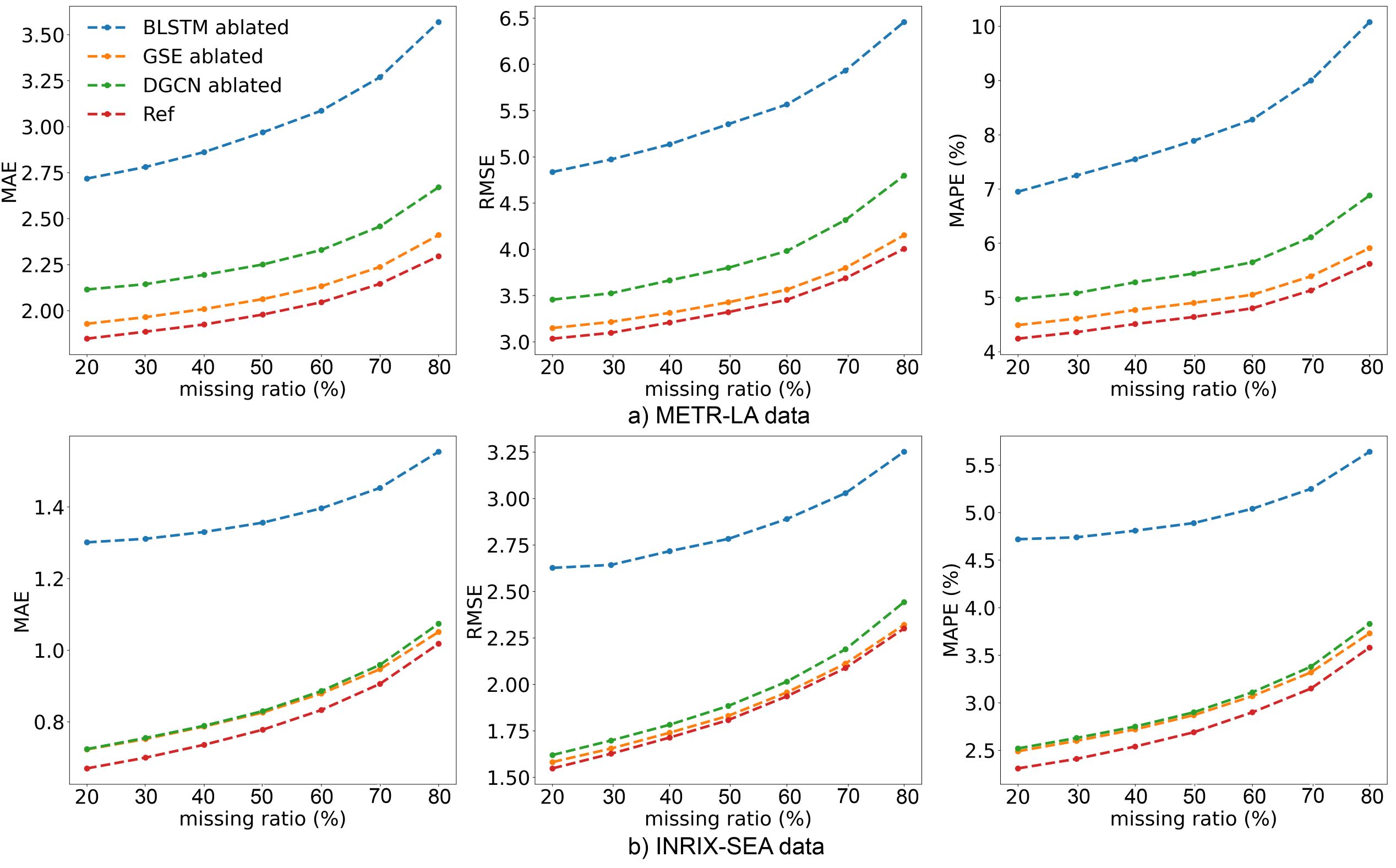}
  \caption{Performance comparison of different modules of DSTGCN for the RM pattern}\label{fig:ablation}
\end{figure}

Fig.~\ref{fig:ablation} shows the results of the ablation analysis. The BLSTM layer is crucial for both datasets, providing a 37.3\%-38.0\% reduction in RMSE for METR-LA data and  29.2\%- 41.1\% reduction in RMSE for INRIX-SEA data. This indicates that temporal correlations play an important role in the traffic data imputation problem for both freeway and urban road networks. 
The GCN layer is also helpful for both datasets. With the GCN layer ablated, the RMSE of METR-LA data increases 13.8\%-19.8\% while the RMSE of INRIX-SEA data increases 4.0\%-6.2\%. Compared with the BLSTM layer, we can find that temporal dependencies contributes more in the traffic data imputation problem than spatial correlations, especially for urban road networks. Additionally, the contribution of the GCN layer to METR-LA data is much more significant than INRIX-SEA data, which agrees with our assumption that spatial correlations on freeway networks are more significant than urban road networks. 
Our proposed graph structure estimation layer is also shown to improve the model performance, which provides a 4.8\%-5.6\% reduction in MAPE for METR-LA data and 4.0\%-7.3\% reduction for INRIX-SEA data. This validates the effectiveness of the GSE layer in capturing dynamic spatial dependencies on traffic networks. 

\subsection{Interpretation Analysis}
This section aims to explore how DSTGCN captures spatiotemporal correlations. To achieve this, we visualize the dynamic and directed transition weights learned from the GSE layer. 
\begin{figure}[!ht]
  \centering
  \includegraphics[width=\textwidth]{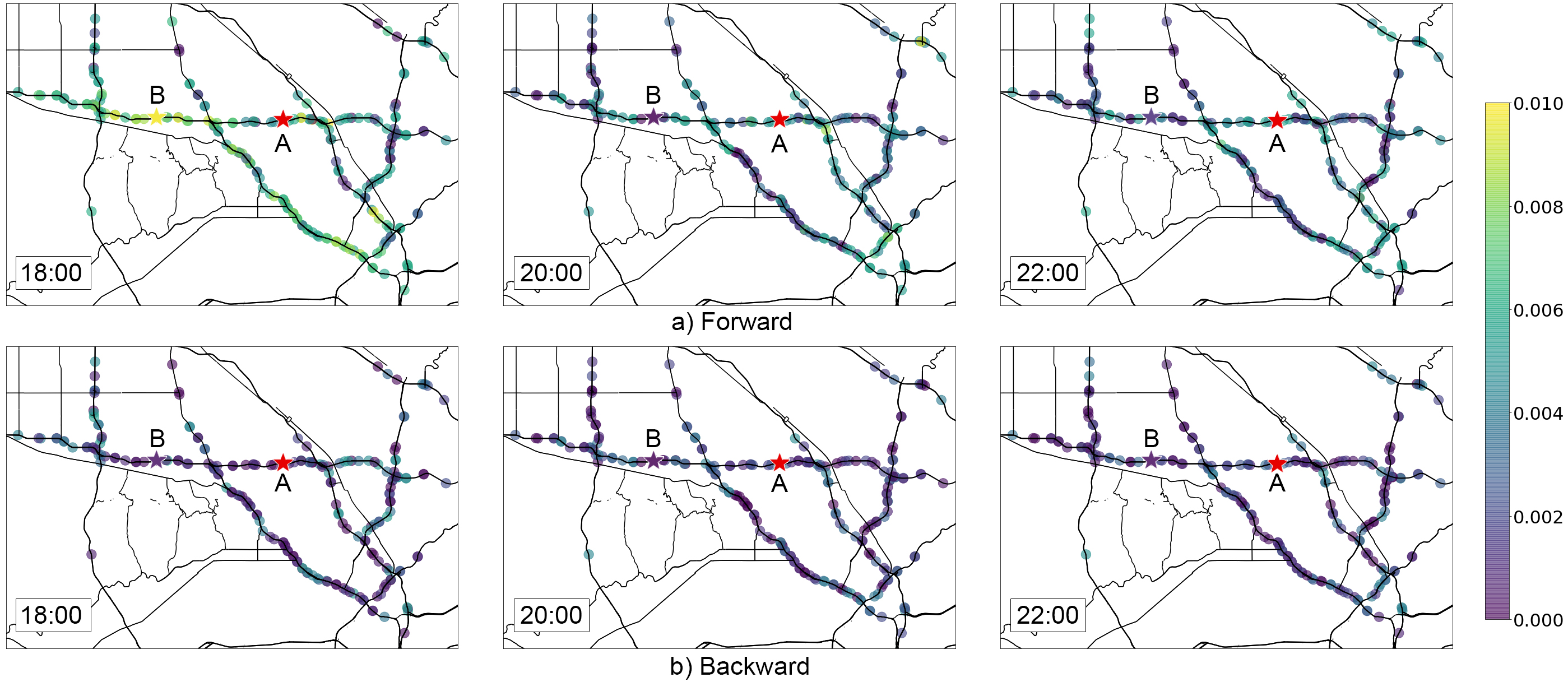}
  \caption{Transition weights between sensor $A$ and the other sensors in METR-LA data from 18:00 to 22:00 on 2012-05-24. Sensor B is marked as an example to highlight the dynamic transition weights between sensors A and B.}\label{fig:dynamic_visual}
\end{figure}

Fig.~\ref{fig:dynamic_visual} displays the learned transition weights between sensor $A$ and the other sensors in METR-LA data across different time. 
Transition weights are normalized using the Softmax function. Each colored dot represents a sensor, and the lighter the color of a dot is, the higher the transition weight between $A$ and the sensor is. Recall that transition weights represent the diffusion likelihood between sensors and higher transition weights indicate stronger spatial correlations. As displayed in Fig.~\ref{fig:dynamic_visual}, spatial correlations are not strictly determined by distance. For example, at 18:00 in the forward direction, sensor $B$ which is not adjacent to $A$ displays high dependencies with sensor $A$. Comparing the transition weights across different time, we can find that sensor $B$ has different correlations with $A$ at different time, indicating that the spatial dependencies are dynamically changing over time. Additionally, even at the same time slot, sensors display different spatial dependencies in the forward and backward direction. This confirms the necessity of capturing spatial dependencies in a directed manner.

\subsection{Comparison with Tensor Factorization}
In recent years, tensor factorization has emerged as a popular approach to traffic data imputation. Generally, tensor factorization methods assume that the multivariate and multidimensional time series can be characterized by a low-rank structure with shared latent factors \cite{chen2021bayesian}. As their underlying model structure and assumptions are distinctly differently from deep learning models discussed in prior sections, it is valuable to formally compare the performance and behavior of different model families. In this section, we compare the performance of our proposed deep learning model with a representative tensor factorization model under two different conditions, based on whether complete training data is available or not. Specifically, we compare DSTGCN with BGCP \cite{chen2019bayesian} as an example of tensor factorization models.

In prior sections, we use complete historical data for model training. It is expected that, when sufficient complete data is provided, deep learning models have a clear advantage over the other methods: first, they can capture the complex spatiotemporal dependencies hidden in the road network, and second, they can be pre-trained for online application. Note that the missing ratio of traffic data can often vary over time. For example, when new traffic sensors are installed, the data is close to complete, but its missing ratio may increase as the sensors degrade over time. In such cases, we may take advantage of the (near)complete data in the beginning to train a deep learning model for imputation later on.
However, it is not always possible to obtain sufficient complete data to train deep learning models. In cases when traffic sensors are damaged for a long period (e.g., several months) or few vehicles are equipped with GPS devices, only incomplete traffic data is available. In such cases, tensor factorization models may offer more robust results, as they directly learn the low-rank structure from the incomplete data without the need for separate model training. 

Because the two models to compare are fitted in very different ways, we design our experiment settings as follows. When there is complete training data, we train DSTGCN using the method described in Section~\ref{exp: settings}. For BGCP, the entire data set (concatenated training, validation and test set) is used as input and the recovery results are generated at once. When only incomplete data is provided, we assume the entire data set has uniform missing distributions. For example, in the RM 60\% scenario, both the training set and the test set have 60\% randomly missing values. For the training of DSTGCN, the training set with masked data is fed to Alg.~\ref{alg:train_sample} to generate training samples. It should be noted that the masked data is kept as ``missing'' during the whole training process, so that DSTGCN can only use the observed information for model training. For BGCP, the entire data set with the same missing distribution is used as input. 
The tensor rank for the BGCP model is set as 40 for METR-LA data and 45 for INRIX-SEA data through extensive experiments.

\begin{figure}[ht!]
  \centering
  \includegraphics[width=\textwidth]{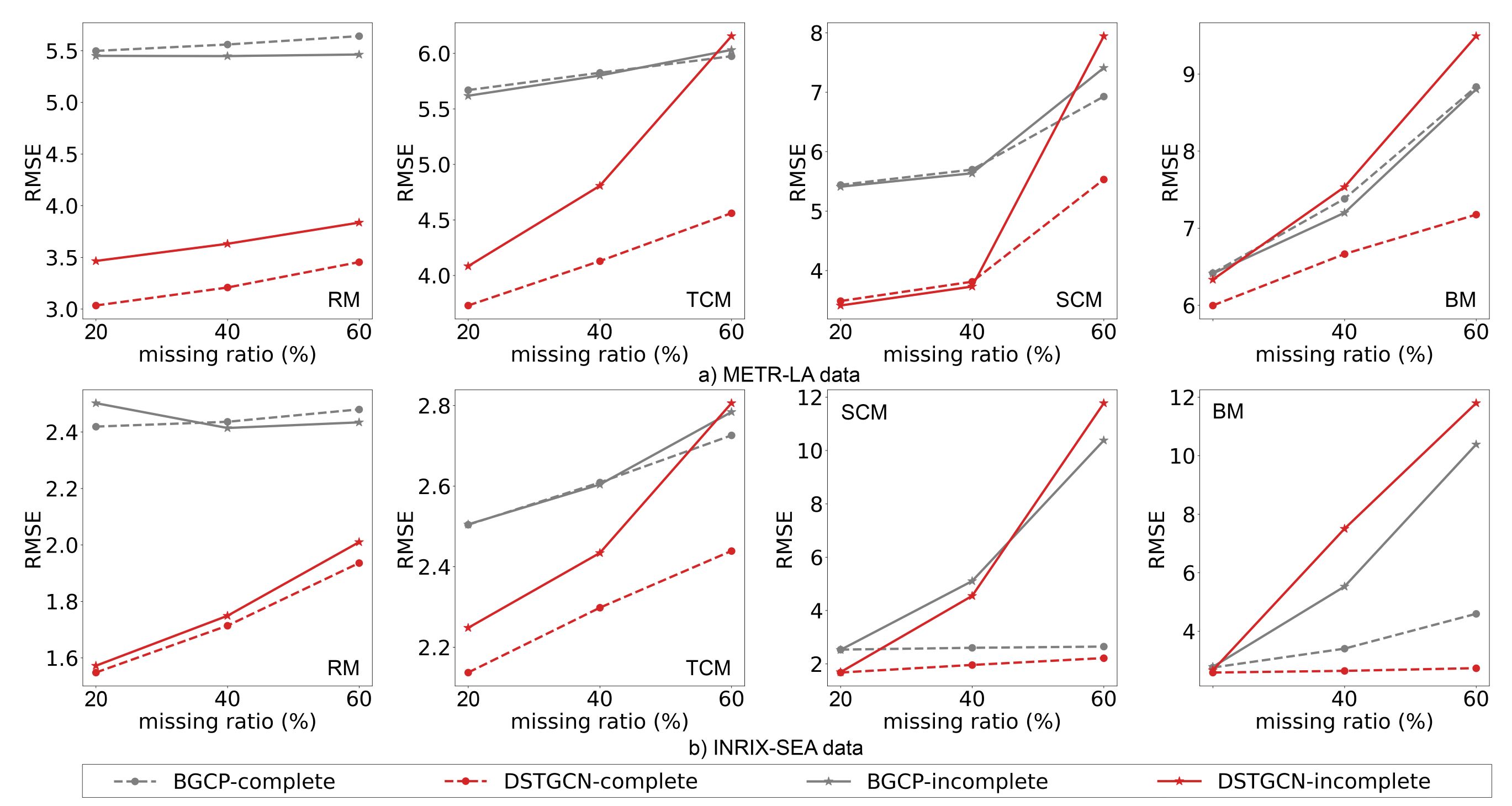}
  \caption{Performance comparison of DSTGCN and BGCP with complete or incomplete training data}\label{fig:DL_vs_TC}
\end{figure}

Fig.~\ref{fig:DL_vs_TC} displays the imputation performance (RMSE) of DSTGCN and BGCP with missing ratios from 20\% to 60\% using complete or incomplete training data. It can be found that when complete training data is provided, DSTGCN outperforms BGCP by a notable margin in all kinds of missing patterns and missing ratios, indicating the effectiveness of deep learning models with sufficient training data. In the absence of complete training data, DSTGCN still achieves superior performance to BGCP in the RM pattern for all missing ratios. In TCM and SCM, DSTGCN performs better than BGCP with missing ratios from 20\% to 40\%, whereas BGCP shows better performance when the missing ratio is 60\%. In BM, DSTGCN and BGCP provides similar performance when the missing ratio is low, whereas BGCP outperforms DSTGCN and the gap is enlarged as the missing ratio gets higher. This result shows the advantage of tensor factorization methods in dealing with incomplete training data with complex missing patterns. This is reasonable as deep learning models are more data-driven while tensor factorization does not require a large amount of high-quality training data.
In summary, deep learning models and tensor factorization models are applicable to different cases: deep learning models perform better when complete training data is provided or the missing pattern is relatively simple, whereas tensor factorization methods are preferred in complex missing scenarios without sufficient training data.

\section{Conclusion}

Missing values in traffic data are an inevitable problem of intelligent transportation systems. Despite many studies to address the issue, there exist two important limitations: first, existing approaches fail to capture the dynamically changing spatial and temporal dependencies in traffic networks; second, most prior studies are based on randomly missing patterns and other complex missing patterns are less considered. To fill these research gaps, this paper introduces a novel deep learning-based framework called DSTGCN to reconstruct missing traffic data. The proposed method comprises several spatiotemporal blocks to capture spatiotemporal information coherently. Each block contains a bidirectional recurrent layer to capture temporal correlations and a diffusion graph convolution layer to capture spatial correlations. Moreover, we introduce a graph structure estimation module to model the dynamic spatial dependencies evolving over time. Extensive experiments are conducted to compare our proposed model with several state-of-the-art deep learning models in four types of missing patterns using two public traffic speed datasets. The results show that our proposed model achieves superior performance to existing methods in different missing patterns and provides robust results with a wide range of missing ratios. Ablation analysis is done to validate the contribution of different model components, and the interpretability of model results are illustrated. In addition, we compare our proposed model with a tensor factorization method under different training data availability. It is found that the deep learning model significantly outperforms the tensor factorization method with high-quality training data whereas tensor factorization is more suitable to deal with incomplete data with complex missing patterns.

There are several directions for future works. First, the DSTGCN model can be adapted to other spatiotemporal problems, such as traffic prediction, route planning, etc. Second, for complex missing scenarios without sufficient training data, deep learning models fail to work well. It would be meaningful to apply transfer learning techniques in the traffic data imputation problem, so that we can leverage models trained in data-rich environment for scenarios without adequate training data. Third, existing imputation researches are usually based on traffic data from either loop sensors or GPS probes. Loop sensors provide traffic data with high sampling rates yet limited spatial coverage, whereas GPS probes cover a wide spatial range with low temporal resolutions. By considering traffic data from GPS probes and loop sensors simultaneously, it is possible to recover traffic data with high spatial and temporal resolutions.


\appendix

\section{Methods for Missing Pattern Generation}\label{missing_generation}

Suppose the missing ratio is $r$ and the imputation time window is $[1, T]$, we generate the missing patterns using the following methods:

\textbf{RM}: A random integer $a$ within range $[0, 1)$ is generated for each data point. If $a < r$, the data point is masked.

\textbf{TCM}: For each node in the traffic network, randomly choose a time point $t$ within $[1, T]$. If $t \leq T - \lfloor T * r \rfloor$,  data points within $[t, t+\lfloor T * r \rfloor]$ are masked. If not, data points within range $[1, \lfloor T * r \rfloor - (T - t))$ and $[t, T]$ are masked.

\textbf{SCM}: For each time slot in the imputation window, randomly choose a node $v$ in the traffic network. If the traffic network contains a set of sensors, the $\lfloor N * r \rfloor$ sensors closest to $v$ are searched and masked. If the traffic network contains a set of road links, we use the Breadth-first Search Algorithm to iterate over the road network from $v$ and the first $\lfloor N * r \rfloor$ road links to be discovered are masked.

\textbf{BM}: BM contains missing data points which are both temporally and spatially correlated. We use Alg.~\ref{alg:BM} to generate the BM pattern.
\begin{algorithm}[ht]
\SetAlgoLined
\SetKwInOut{KwIn}{Input}
\SetKwInOut{KwOut}{Output}
\KwIn{missing ratio $r$, imputation window $[1, T]$, the input data matrix $S$}
 $i=1$\;
 \While{$i \leq T$}{
  Randomly generate an integer $a$ within range $[1, T-i]$\;
  Randomly choose a node $v$ from the traffic network\;
  Select the $\lfloor N * r \rfloor$ closest nodes $N_v$ to $v$\;
  Mask the data points of $N_v$ within time range$[i, i+a]$ in $S$\;
  $i = i + a$\;
  }
 \caption{Generating the BM pattern on a data matrix $S$}\label{alg:BM}
\end{algorithm}

\section{The imputation performance of DAE for SCM and BM with INRIX-SEA data}\label{DAE_performance}

\begin{figure}[!ht]
  \centering
  \includegraphics[width=\textwidth]{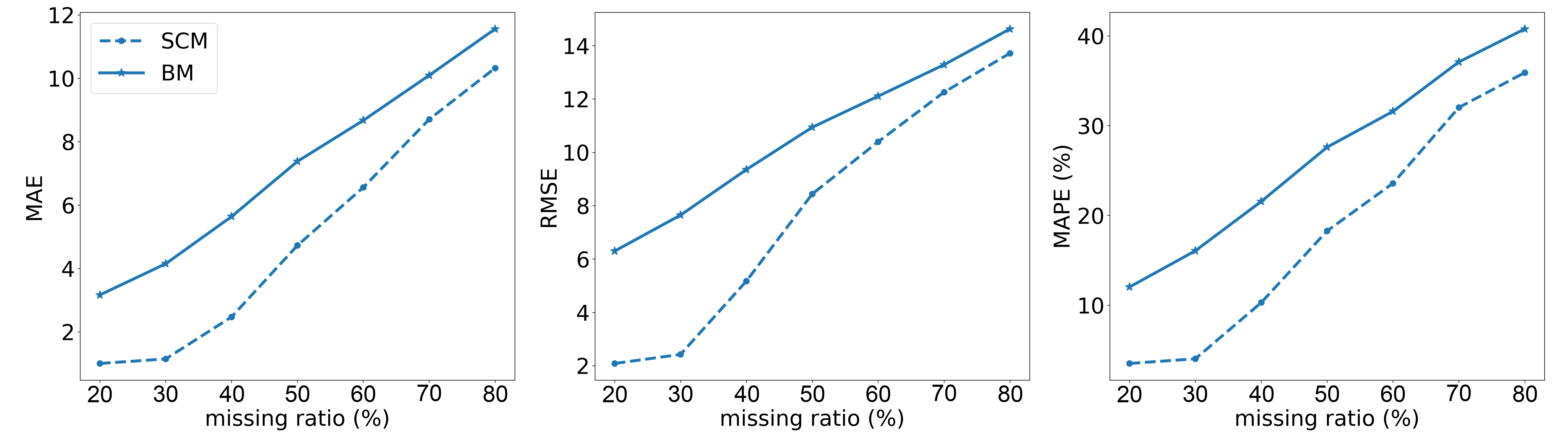}
  \caption{The performance of DAE for SCM and BM with INRIX-SEA data}\label{fig:DAE_res}
\end{figure}

The imputation performance of DAE in SCM and BM patterns for INRIX-SEA data is displayed in Fig.~\ref{fig:DAE_res}.

\bibliographystyle{unsrt}  
\bibliography{ref.bib}

\end{document}